\documentclass[10pt,conference]{IEEEtran}
\usepackage{cite}
\usepackage{amsmath,amssymb,amsfonts}
\usepackage[linesnumbered,ruled]{algorithm2e}
\SetKwComment{Comment}{/* }{ */}

\usepackage{balance}

\usepackage{graphicx}
\usepackage{textcomp}
\usepackage{xcolor}
\usepackage{comment}
\usepackage{booktabs}
\usepackage{multirow}
\usepackage{xspace}
\usepackage{tcolorbox}

\usepackage{array}

\def\BibTeX{{\rm B\kern-.05em{\sc i\kern-.025em b}\kern-.08em
    T\kern-.1667em\lower.7ex\hbox{E}\kern-.125emX}}

\newcommand{\tool}{CDSeer}
\newcommand{\toolS}{CDSeer\xspace}
\newcommand{\PreserveBackslash}[1]{\let\temp=\\#1\let\\=\temp}
\newcolumntype{C}[1]{>{\PreserveBackslash\centering}p{#1}}
\newcolumntype{R}[1]{>{\PreserveBackslash\raggedleft}p{#1}}
\newcolumntype{L}[1]{>{\PreserveBackslash\raggedright}p{#1}}

\newcolumntype{?}{!{\vrule width1pt}} 
\def\BibTeX{{\rm B\kern-.05em{\sc i\kern-.025em b}\kern-.08em
    T\kern-.1667em\lower.7ex\hbox{E}\kern-.125emX}}
    
\newcommand{\pa}[1]{\noindent\textbf{#1}}
\newcommand{\RQOneAlt}{RQ1-SOTA: What are the performance and limitations of the state-of-the-art concept drift detection approaches on open-source and proprietary datasets?}

\newcommand{\RQTwo}{RQ2-\tool: How does \toolS perform compared to the SOTA methods in concept drift detection?}
\newcommand{\RQThree}{RQ3-Generality: Can \toolS 
 detect concept drift in pre-trained models?}

\newcommand{\rqboxc}[1]{\begin{tcolorbox}[left=4pt,right=4pt,top=4pt,bottom=4pt,colback=gray!5,colframe=gray!40!black,before skip=4pt,after skip=4pt]#1\end{tcolorbox}}

\begin{document}

\title{Time to Retrain? Detecting Concept Drifts in Machine Learning Systems}
\author{\IEEEauthorblockN{Tri Minh Triet Pham}
\IEEEauthorblockA{Concordia University\\
Montreal, Canada\\
p\_triet@encs.concordia.ca}
\and
\IEEEauthorblockN{Karthikeyan Premkumar}
\IEEEauthorblockA{Ericsson\\
Montreal, Canada\\
karthikeyan.premkumar@ericsson.com}
\and
\IEEEauthorblockN{Mohamed Naili}
\IEEEauthorblockA{Ericsson\\
Montreal, Canada\\
mohamed.naili@ericsson.com}
\and
\IEEEauthorblockN{Jinqiu Yang}
\IEEEauthorblockA{Concordia University\\
Montreal, Canada\\
jinqiu.yang@concordia.ca}
}

\maketitle

\begin{abstract}
With the boom of machine learning (ML) techniques, software practitioners are building ML systems to process massive volumes of streaming data for diverse software engineering tasks, such as failure prediction in AIOps. Trained on historical data, these models can suffer performance degradation due to concept drift, where the inter-relationship between features and labels (concepts) changes between training and production. Therefore, concept drift detection is essential to monitor deployed ML models and retrain them as needed. 

In this work, we explore applying state-of-the-art (SOTA) semi-supervised concept drift detection techniques on synthetic and real-world datasets in an industrial setting, which requires minimal manual labeling and maximal model compatibility. We find that current SOTA methods not only require significant labeling effort but also are model-specific. To address these limitations, we propose \tool, a novel model-agnostic technique to detect concept drift.

Our evaluation shows that \toolS outperforms the SOTA in precision and recall while requiring significantly less manual labeling. We demonstrate \tool's effectiveness at concept drift detection by evaluating it on eight datasets from different domains and use cases. Internal deployment of~\toolS on a proprietary industrial dataset shows a 57.1\% improvement in precision while using 99\% fewer labels compared to the SOTA method. Its performance is also comparable to the supervised (fully labeled) concept drift detection method. The improved performance and ease of adoption make \toolS valuable in enhancing the reliability of ML systems.
\end{abstract}


\section{Introduction}

Modern systems generate huge amounts of data and are expected to reach 463 exabytes by 2025~\cite{muchdata} where a significant portion of the data consists of time series data streams.
The data can come from sensor recordings and signals from internet-of-things (IoT) and autonomous driving systems (ADS)~\cite{muchdata}; or alerts, logs, and measurements from software systems~\cite{Chen2019OutagePA, ElSayed2017LearningFF, Xue2018SpatialTemporalPM, He2018IdentifyingIS}.
As these systems automatically produce data at high volume and velocity, posing a significant challenge to manual processing capabilities, practitioners develop machine learning (ML) systems powered by ML models to perform critical tasks such as predicting software failures~\cite{Lyu2023OnTM}.

One open challenge to the reliability of deployed ML systems is that the ML models inevitably become outdated and experience degraded performance on new data due to the constantly changing nature of real-world environments and data distributions.
Such differences can be categorized as data or concept drift. Data drift occurs when production data follows a different distribution from training data or is unseen by the model during training.
Conversely, concept drift happens where the data features remain the same, but the relationship between the features and the class (concepts) changes.
Hence, it is important to continuously test and monitor a deployed ML system for both data and concept drift. Once a drift is detected, maintainers of the ML system should be alerted that the deployed model is obsolete and it is time to retrain.


Current concept drift detectors (CDDs) are classified into three types: supervised CDDs, which only use labeled data for detection; unsupervised CDDs, which only use unlabeled data; and semi-supervised CDDs, which use a mix of unlabeled and labeled data~\cite{BayramConceptDriftSurvey, Haque2016ECHO, Pinage2019ADD}.
Supervised CDDs are impractical as manual labeling true labels for all data is time-consuming and costly, especially when a domain expert is needed. 
Unsupervised CDDs~\cite{Gemaque2020AnOO} rely on detecting data drifts to identify concept drifts; therefore, it cannot reliably detect concept drift in the absence of data drift.
Instead, semi-supervised CDDs are preferred for their practicality, as they significantly reduce manual labeling effort~\cite{Haque2016ECHO, Pinage2019ADD}. However, this reduction in effort can be influenced by the frequency and volume of data a system receives. In real-world systems, data can arrive in large volumes and at unpredictable rates, making it challenging to accurately identify drift without overwhelming computational resources or manually labeling ground truth (GT)~\cite{Haque2016ECHO, Pinage2019ADD}. Additionally, the varying quality and representativeness of incoming data can impact the effectiveness of semi-supervised methods, necessitating strategies to adapt dynamically to changing data streams.


Particularly, we applied ECHO~\cite{Haque2016ECHO}, the SOTA semi-supervised CDD, on various datasets and models, including one internally in Ericsson's setting, one dataset from Google on predicting job failure~\cite{wilkes2020google}, and four datasets from prior works. Our experiment shows that the SOTA technique requires a non-trivial amount of manual labels up to 100\%, calling for better concept drift detection techniques. 
When we discuss these results on real datasets and their implications if the SOTA CDDs are applied in an industrial setting, we find the limitations make the SOTA CDDs unsuitable for real-world deployment. Considering these results in real datasets and their implications for applying state-of-the-art (SOTA) CDDs in an industrial setting, we find that the limitations make these SOTA CDDs unsuitable for real-world deployment.

We summarize three key industry concerns that SOTA techniques fail to address.
\begin{itemize}
    \item \textbf{Excessive labeling effort.} The SOTA semi-supervised CDDs require an average of 5\%, and up to 100\% of the time series to be labeled~\cite{Sriwatanasakdi2017ConceptDD}. While this represents a significant improvement over supervised CDDs, it remains challenging to achieve in production.
    
    \textbf{Industry Need.}
    Experts should be required to label as few data points as possible, and the labeled points should be as useful as possible, i.e., minimizing manual labeling while maximizing accuracy in detecting concept drift.
    
    \item \textbf{Lack of flexibility in selecting which data points require labeling.} The SOTA CDDs request manual labels of specific data points individually and on-demand based on dataset and model metrics, i.e., cannot tell \textit{when} and \textit{which} labels are needed beforehand. This limits the ability to plan labeling efforts.
    
    \pa{Industry Need.}
    Experts prefer to label multiple data points in each session at their convenience.
    Also, they prefer to select as few points to label as possible, preferably points that cost them the least labeling effort.

    \item \textbf{Lack of generality for different ML models.} SOTA CDDs are optimized for specific model architectures and cannot be generalized to different model architectures.

    \pa{Industry Need.}
    There is much need for black-box (model architecture-agnostic) CDD since ML models can already be trained, and it is difficult to convince users to adopt a new model, especially if the current one is optimized for their use case.
\end{itemize}


To address the challenges above, we propose \toolS, a novel semi-supervised CDD. \toolS significantly reduces manual labeling workload and is both model- and distribution-agnostic. \toolS works with ML models of various architectures and is independent of the underlying data distribution.  
CDSeer's innovations are as follows.
\begin{enumerate}
    \item \textbf{Model- and data distribution-agnostic predictable sampling for manual labeling}. 
    \toolS employs a novel sampling process that eliminates reliance on model and data specifics, enabling it to detect concept drift where SOTA methods may fail. This process is also predictable in terms of label requests and their frequency, allowing for better planning of labeling efforts.
    \item \textbf{Semi-supervised learning using an inspector model to provide pseudo-GT labels for concept drift detection}. We propose a novel method that leverages an inspector model to generate pseudo labels to replace true labels as the GT for concept drift detection. This significantly reduces the number of required labels while maintaining competitive performance compared to SOTA.
    \item \textbf{Model-agnostic concept drift detection}. \toolS can monitor ML models regardless of their architecture. In contrast, SOTA CDDs~\cite{Pinage2019ADD, Haque2016ECHO} rely on ensemble classifiers for both prediction and concept drift detection, making them unsuitable for pre-trained industry models.
\end{enumerate}

Our contributions include (1) a novel semi-supervised CDD that outperforms the SOTA techniques and satisfies industry constraints, and (2) a comprehensive evaluation of \toolS. 
We evaluate~\toolS on publicly available synthetic and real datasets, as well as a proprietary dataset internal to Ericsson. The study on the proprietary dataset shows that \toolS achieves comparable performance improvements in model accuracy to the supervised method and outperforms the SOTA semi-supervised method while using significantly fewer labeled data points.

\pa{Relevance to Software Engineering in Practice.} First, serving ML systems and detecting performance degradation over time (such as model obsolescence) are active research areas in the Software Engineering community (e.g., \cite{Lyu2023OnTM}). Our work focuses on monitoring and testing deployed ML systems from a \textit{practical} perspective, an important yet overlooked research area. 
Second, research on concept drift techniques can be applied to strengthen downstream SE research. For instance, ML models built to predict production system failures and performance degradation suffer from concept drifts in~\cite{Lyu2023OnTM, xia2023}. 
Third, we propose a novel technique, \toolS, to fill the gap between academic research and software industry practices. \toolS is shown to outperform the SOTA and is deployed in the industry. Our work is well suited for the SEIP track.

\section{Background}
\subsection{Concept Drift and Data Drift}
\label{sec:background_definition}
We notice the interchangeable uses of the terms “concept drift” and "data drift" in literature can cause much confusion about their meaning~\cite{BayramConceptDriftSurvey, Gemaque2020AnOO}. 
Hence for clarity, in this work, we define data drift and concept drift as follows.


Suppose that we have two datasets: the training $D_{training} (X,y)$ and testing $D_{testing} (X,y)$ where $X$ is the input features and $y$ is the target labels. Drifts occur when the input data distribution $P(X)$ or the distribution of the target concept of the data $P(y|X)$ changes.
\begin{itemize}
    \item \textbf{Data drift} happens when $P_{training}(X) \neq P_{testing} (X)$ and $P_{training} (y|X)$ $=$ $P_{testing} (y|X)$.
    Literature\cite{Shimodaira2000ImprovingPI, Gao2007AGF, DELANY2005187, Sugiyama2012MachineLI, Gama2014ASO} also refer to this as ``dataset shift'', ``type I concept drift'', or ``(virtual) concept drift'' (where ``virtual'' is sometimes omitted). This type of drift is commonly detected via distribution monitoring or out-of-distribution detection~\cite{hendrycks2016baseline, Corbire2019AddressingFP, Yang2021GeneralizedOD, pmlr-v162-sun22d}.
    \item \textbf{Concept drift} happens when $P_{training} (X)=P_{testing} (X)$ and $P_{training} (y|X)$ $\neq$ $P_{testing} (y|X)$. In literature, this is referred to as ``type II concept drift'', or ``real concept drift'' (where ``real'' is sometimes omitted confusing it with data drift). This type of concept drift is the main focus of our work.
    \item \textbf{Concept drift and data drift} may happen simultaneously when $P_{training} (X) \neq P_{testing} (X)$ and $P_{training} (y|X)$ $\neq$ $P_{testing} (y|X)$. Literature names this as ``type III concept drift'' or just ``concept drift'' since it is both virtual and real at the same time. This type of drift can be detected by combining concept and data drift detectors for the best performance.
\end{itemize}
Readers can observe the variation in terminology in citations~\cite{MorenoTorres2012AUV, BayramConceptDriftSurvey, Lu2019, Gemaque2020AnOO}. Many methods claim concept drift detection but only monitor $P_{training} (X) \neq P_{testing} (X)$ under the assumption that concept drifts happen simultaneously with data drifts~\cite{DELANY2005187}. Hence, these methods cannot detect concept drift without data drifts. Our work focuses on detecting real concept drift when $P_{training} (y|X) \neq P_{testing} (y|X)$ and $P_{training} (X)=P_{testing} (X)$.
We do not monitor data drift.

\subsection{Semi-Supervised Concept Drift Detection for Machine Learning Systems}

\begin{figure}[ht]
    \centering
\includegraphics[width=\columnwidth,keepaspectratio]{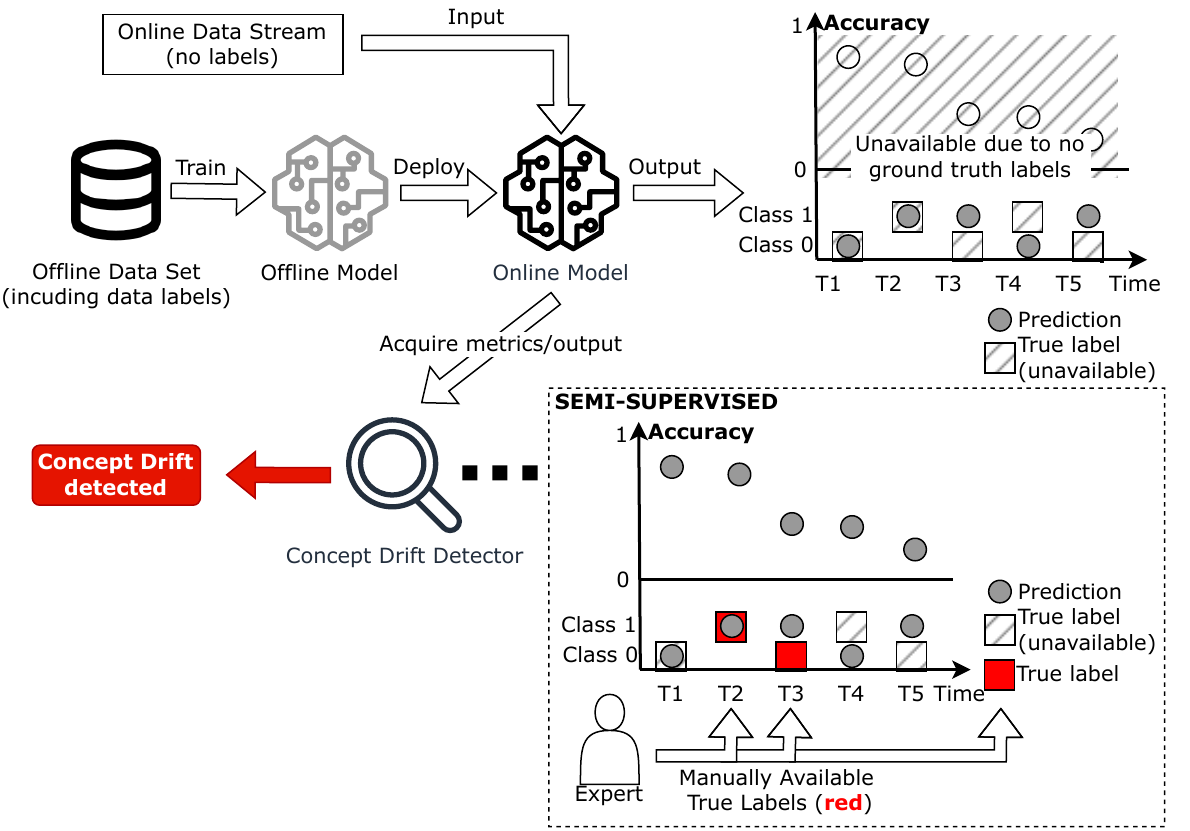}
    \caption{High-level semi-supervised concept drift detection process.}
    \label{fig:background_semi_supervised}
\end{figure}
Figure~\ref{fig:background_semi_supervised} illustrates our context of applying semi-supervised CDD. 
In our scenario, the ML models are first trained offline using labeled data. Then, they are deployed online to classify the time series.
When concept drift happens, it causes accuracy deterioration, such as \texttt{T3}, \texttt{T4}, and \texttt{T5} in the top right chart in Figure~\ref{fig:background_semi_supervised}.
However, in reality, the true labels of the input time series are not available. Thus, it is difficult for maintainers to observe the accuracy deterioration.

CDDs can be categorized into supervised, semi-supervised, and unsupervised methods. 
Supervised approaches require true labels for all data, which is often unrealistic.
In contrast, unsupervised and semi-supervised approaches require true labels for either no data or only a few data points, making them more practical in real-world applications.
Current unsupervised techniques detect concept drift only when it coincides with data drift, rendering them ineffective at detecting standalone concept drifts. The ability to detect concept drifts directly is desirable since there is no guarantee that data and concept drifts happen simultaneously.

Based on our industry needs, we believe semi-supervised approaches are suitable to detect concept drift \textbf{accurately} and \textbf{efficiently} in production. These approaches require minimal manual labeling and can target independent concept drifts.
In a nutshell, our work focuses on proposing a better semi-supervised approach to detect concept drifts as we believe semi-supervised approaches are better aligned with our industry needs. 

\section{Related Works}
\subsection{Concept Drift Detection}
\pa{Supervised concept drift detection.}
Supervised CDDs use the true labels of all data points to detect concept drift.
They either compare statistics of the classifier error rate between the current and reference windows~\cite{adwin, gama2004, pagehinkley1954, Sebastio2017SupportingTP, CieslakChawla2007, Harel2014ConceptDD} or use the intersection of confidence intervals from multiple CDDs to detect concept drift~\cite{Nishida2005ACEAC, Alippi2013JustInTimeCF}.
While these methods are fast and accurate, they require labels for all data points which is unrealistic in production given the velocity of data streams~\cite{Masud2008}.

\pa{(Semi)-supervised concept drift detection.} As a more production-friendly alternative, semi-supervised CDDs solve the problem by requiring only a subset of the data to be labeled. 
Existing methods can do this by using active learning~\cite{Fan2004, Zhu2007, Masud2010}, computational geometry~\cite{Dyer2014}, class imbalance~\cite{Lughofer2016SemiUnsupervised}, classifier confidence~\cite{Haque2016SAND, Haque2016ECHO}, pseudo-error from a classifier ensemble~\cite{Pinage2019ADD}, grid density sampling~\cite{Sethi2016AGD}, ambiguity concerning the decision boundary and neighboring points~\cite{Sriwatanasakdi2017ConceptDD}, etc. to select only important data instances for labeling, reducing the labeling effort.
However, the reliance on specific metrics to sample labels is vulnerable to data distribution and model architecture changes. Also, requiring the original model to follow certain ML architectures significantly limits adoption. 
We look to overcome these limitations.


\pa{Unsupervised concept drift detection.}
Unsupervised CDDs~\cite{Gemaque2020AnOO} detect drifts offline in batch~\cite{Sethi2015DontPF, Sethi2017OnTR, Sethi2018HandlingAC, Liu2018Unsupervised, Li2019FAADAU, Bashir2017AFF, Maletzke2017QuantificationID, Maletzke2018OnTN, Maletzke2018CombiningIS, Costa2018ADD} or online by instances~\cite{Tsymbal2008DynamicIO, Reis2016FastUO, Koh2016CDTDSCD, Kim2017AnEC, Mustafa2017UnsupervisedDE, Mello2019OnLG, Cerqueira2021STUDDAS, gozuacik2019, Baier2021DetectingCD, Kingetsu2021BornAgainDB}.
These methods detect concept drift by monitoring $P_{training} (X) \neq P_{testing} (X)$ under the assumption that changes in the data distribution and the posterior probability distributions often happen simultaneously~\cite{DELANY2005187}.
Hence, unsupervised methods cannot detect concept drift when data drift is absent (Subsection~\ref{sec:background_definition}).
~\toolS aims to detect real concept drift even when the input data distribution stays the same.


\subsection{Testing Machine Learning Systems}
Testing ML systems is an important research area in SE~\cite{ml_testing_survey}, of which, black-box testing, especially of deployed models, is of particular interest due to being source code/architecture independent and generalizable to different systems.

\pa{Artificial Intelligence for IT Operations (AIOps)}
AIOps leverage ML models 
for tasks such as anomaly detection and failure prediction~\cite{Lim2014IdentifyingRA, Ros2015CatchingFO, Botezatu2016PredictingDR, ElSayed2017LearningFF,Mahdisoltani2017ProactiveEP, Xu2018ImprovingSA, Lin2018PredictingNF, He2018IdentifyingIS, Chen2019OutagePA, Li2020TOSEM}
using logs, measurements, and spatial information.
These models are highly susceptible to data evolution, which impairs their performance and stability.
Current mitigation strategies involve periodically retraining models~\cite{Lin2018PredictingNF, Li2020TOSEM} or retraining based on alarms from supervised CDDs~\cite{Lyu2023OnTM}, both of which are costly in manual efforts due to the extensive labeling required.
In this work, we explore~\toolS's concept drift detection for AIOps to reduce cost.

\pa{Model Robustness and Relevance Testing.}
Model testing detects mismatches between the model and data~\cite{Zhang2019PerturbedMV, WerpachowskiOverfitting} and ensures trained models' resistance to data noise~\cite{Ruan2019GlobalRE, Mangal2019RobustnessON, Banerjee2019TowardsBA}.
However, these techniques test pre-trained models offline and more online techniques are needed~\cite{ml_testing_survey}.
\toolS detect concept drift online to test the robustness and relevance of ML models.
Thus ~\toolS can benefit the utilization and deployment of ML models for SE tasks. In this work, we applied \toolS to an internal dataset for anomaly detection in router systems and the Google Cluster Trace dataset for job failure prediction.

\section{\tool: Learning-based Concept Drift Detection}
\label{sec:method_overview}
In this section, we first summarize our approach at a high level.
Then, we describe the framework step-by-step, with design choices and implementation details.
\subsection{An Overview of \tool}
\begin{figure}[ht]
    \centering
    \includegraphics[width=\columnwidth,keepaspectratio]{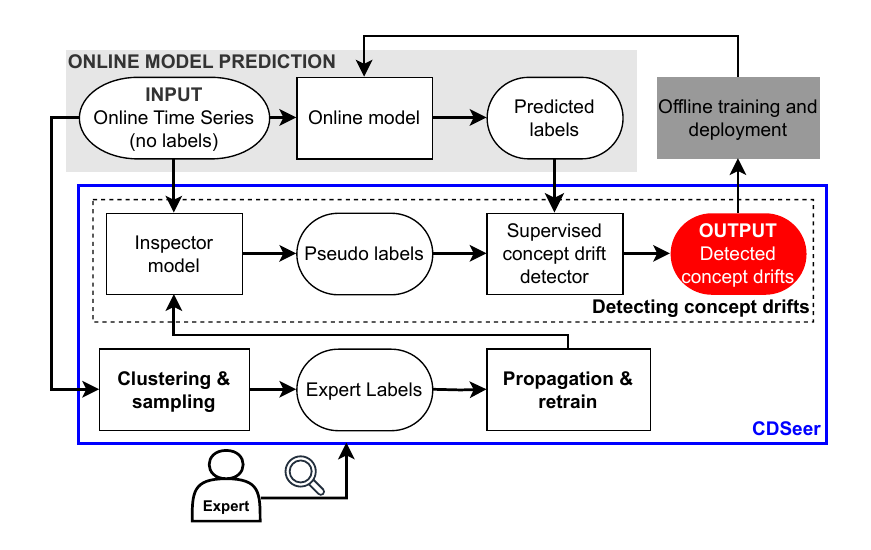}
    \caption{An overview of our proposed framework \tool.}
    \label{fig:framework_overview}
\end{figure}
Figure~\ref{fig:framework_overview} shows an overview of our proposed concept drift detection technique namely \tool.
In this figure, the components are separated into two parts.
The grey box \textbf{Online Model Prediction} shows a generic data stream classification process, where the online classifier model predicts the labels of incoming streaming data points.
The blue box called \textbf{\toolS} summarizes our framework at a high level.

At a high level, \toolS works entirely online as follows. As new data points arrive, the online model predicts labels in parallel with the inspector model predicting pseudo-labels. Concept drift is detected by a supervised CDD that monitors the distribution of classification error between the labels and pseudo-labels.
The inspector model is trained as follows: the window of the latest data points is continuously clustered. 
From these clusters, a stratified sample is selected for labeling by experts and stored in memory.
Via semi-supervised learning, the true labels in memory are propagated back to the window of the latest data points, where these newly generated labels and their features are used to update the inspector model. 
Finally, while this is not the main focus of our experiments, we follow convention and retrain the online model when concept drift is detected using true labeled data.

\subsection{Offline Training of the Inspector Model}
During online deployment, the inspector model is retrained automatically (as described in later sections). However, before online deployment, the inspector model is trained offline using (a sample of) the online model's training dataset. 
In this state, it cannot detect concept drifts until retrained online.

\subsection{Concept Drift Detection}
\begin{algorithm}
    \caption{Proposed Concept Drift Detection \textit{\tool}}\label{alg:framework_algo}
    \KwData{trained $OnlineModel$, $InspectorModel$, and stream of unlabeled data points $X$}
    \KwResult{the points with concept drift}
    \For{each $x$ in $X$}{
        \tcc{online and insp. models predict labels}
        $\hat{y} \gets OnlineModel(x)$\;
        $\hat{y}_{insp} \gets InspectorModel(x)$\;
        \textbf{If} $\hat{y}=\hat{y}_{insp}$ \textbf{then} $error \gets 1$\ \textbf{else} $error \gets 0$ \textbf{end}\;
        $CDD \gets error$\tcp*[l]{CDD monitors error}
        \If {$CDD$ raise alarm}{
            record $x$ index and notify maintainers\;
        }
        $W \gets$ the latest $w$ unlabeled points including x\;
        $clusters \gets$ cluster unlabeled points in $W$\;
        $candidates \gets$ stratified sample from $clusters$\;
        $true\_labels \gets$ labeled $candidates$ from experts\;
        \If {$true\_labels$ is not empty}{
            $M \gets true\_labels$ and flush oldest points exceeding pre-defined fixed size\;
            \tcc{each point in $W$ is assigned a label via $LabelSpreading$ from true labels in $M$}
            $(W, TrueLabels) \gets$ spread labels from $M$ to $W$\;
            \textbf{InspectorModel} $\gets$ trainModel$(W, TrueLabels)$\;
        }
    }
\end{algorithm}

Each incoming unlabeled data point ($x$) in the time series is classified by both the online and the inspector models to create two labels: $\hat{y}$ and $\hat{y}_{insp}$ from respectively.
The $error$ between the labels is 0 if they are the same and 1 otherwise. 
\begin{equation}
error_i = 
\begin{cases}
    0,              & \text{if } \hat{y}=\hat{y}_{insp,i}\\
    1,              & \text{otherwise}
\end{cases}
\forall i \in \mathcal{D}
\end{equation}

We utilize a supervised CDD to monitor the distribution of the $error$s rate and alert when it crosses a threshold, signaling the online and inspector models are trained using different concepts.
For this purpose, the CDD can be chosen from well-known CDDs such as the Drift Detection Method~\cite{gama2004}, Page-Hinkley test (PHT)~\cite{pagehinkley1954}, etc.
The CDD monitors the aggregated $error$ rate between the online and inspector models for increases surpassing a predefined threshold, which triggers a concept drift alarm between the online and the inspector models. Then, the framework records the corresponding instance and proceeds to the next data point.

\subsection{Clustering and Sampling for Label Requisition}
\subsubsection{Clustering Data Stream}
From the current data point $x$, we group the previous $w$ consecutive points including $x$ into a window $W$ which is then clustered via DBSCAN~\cite{dbscan, skmultiflow}. DBSCAN expands from high-density core samples (measured by Euclidean distances) to form clusters, which removes the need to specify the number of clusters upfront. 
The cores are computed from the pre-defined minimum number of points in each cluster ($minPts$) and the maximum distance ($\epsilon$).
The $minPts$ is calculated as $\frac{w}{100}$ to ensure that clusters are sufficiently large and no more than 1\% of the dataset requires labeling. 
We estimate $\epsilon$ using a simplified version of knee detection~\cite{Satopaa2011FindingA}. We compute the Euclidean distance between every point in the training dataset and its nearest neighbor and plot them in ascending order. The point with the largest curvature is chosen as the $\epsilon$ since most points would have smaller distances to their neighbor and can form high-density clusters for DBSCAN.
\begin{equation}
clusters = DBSCAN(W, Euclidean, \epsilon, \frac{w}{100})
\end{equation}

However, when there are no high-density cores due to randomly generated features, DBSCAN can form too many or only a single cluster. We address this via K-means clustering~\cite{kmeansoreilly, skmultiflow}. 
The initial centroids are randomized with a count of $\frac{w}{100}$ to ensure at most 1\% of the dataset requires labels.
The K-means algorithm computes the Euclidean distance and assigns points closest to the centroids to their respective clusters where the mean of each cluster becomes the new centroids. This process is repeated until convergence yielding the clusters surrounding the centroids.
\begin{equation}
clusters = KMeans(W, Euclidean, \frac{w}{100})
\end{equation}

When inputs are images or text, we also leverage K-means for clustering after pre-processing the input. For images, the features are the pixels of the image with some caveats.
First, the pixels are in grayscale and normalized to have a value between 0 and 1.
Second, the size of each image needs to be very small (no more than 28x28 pixels).
Datasets without these characteristics may need feature reduction and grayscale conversion.
We can then compute distance treating each pixel as a feature.
When the inputs are texts, we follow the original pre-processing steps to clean, trim, vectorize, and tokenize the input texts~\cite{tf_text_classification}. This results in the features being fixed-sized sparse feature vectors with numbered values representing the encoded vocabulary that can be used to compute distances representing similarities between texts.

\subsubsection{Sampling from Clusters}
From the previous step, $W$ is clustered into several $clusters$ of points. Then, one point from each cluster is sampled for labeling. The points within the same cluster are meant to be interchangeable.
\subsubsection{Labeling by Experts}
Then, experts will label the sample similar to how they label requests in SOTA methods, with two improvements: (1) the expert can label only a subset of the sample, and (2) the expert can exchange the sampled point with another from the same cluster if desired.

\pa{Rationale behind clustering}.
We arrive at clustering from our observations of how the current SOTA works.
SOTA semi-supervised methods request labels based on metrics of the target data/model. For example, ECHO selects data points for labeling based on the classifier model's confidence.
Depending on the metric, SOTA may request labels for 0-100\% of the testing data, which can cause the detection process to fail or be highly inefficient.
On the other hand, simple random label requests are inefficient and costly, as we risk labeling unnecessary points.
Hence, clustering is a compromise that directs the manual labeling process to the most representative points in each window via stratified sampling.

\pa{Discussion on how \toolS addresses industry needs}.
Clustering is meant to address several industry needs.
First, data points are selected so experts can label multiple points in a single session.
Second, the data points summarize the current data, preventing efforts spent labeling highly similar points.
Third, not all selected data points need to be labeled, and data points within the same strata are interchangeable, allowing experts to label the points that require the least effort.
This approach offers greater flexibility needed in the industry compared to SOTA.

\subsection{Propagation and Retraining the Inspector}
All manually labeled data points are managed by memory $M$. $M$ is a fixed-sized queue of the newest data points with true labels available where the oldest points are automatically removed when the pre-defined size of $M$ is exceeded.
Since the true labels (in $M$) are scarce, we increase their number via
semi-supervised learning.
To do this, we combine $M$ and $W$ and apply semi-supervised learning so true labels spread from $M$ to $W$ to get $(W, TrueLabels)$ where all points have labels.
We select $LabelSpreading$ in sklearn~\cite{Zhou2003LearningWL}, a graph inference-based algorithm that is robust to noises and can accurately spread very few labeled points to a large number of unlabeled points. This method constructs a complete weighted graph between all data points (labeled and unlabeled) where the weights are the distance between the points. Labels would spread from labeled to unlabeled data points iteratively until convergence where a higher weight slows spreading. If an unlabeled data point is assigned multiple different labels, the most frequently assigned label is selected as the final label for the point.
Then, we retrain $InspectorModel$ using $(W, TrueLabels)$. 

Our intuition is that when concept drift occurs, $M$ would contain a mix of both old and new concepts. When labels in $M$ are spread to $W$, it results in an inspector model that is trained on mixed concepts.
We theorize that predictions from such an inspector model would increase the $error$s between the online and inspector's predictions, triggering CDD alarms signaling the original concept has shifted.
This way, the inspector's predictions can be low-cost substitutes for the true labels. Our reasoning behind utilizing the window $W$ is two-fold:
First, the true labels received are too few to train the inspector, hence, label spreading is used. The key for label spreading to work is the assumption that points belonging to the same clusters are likely to have the same labels.
Second, using the window $W$ of the most recent points is likely to reflect the current concept and preserve the characteristics of the data stream.
This way, supervised CDD is converted to semi-supervised, significantly reducing the number of labeled data needed.


\section{Evaluation Setup}
\label{sec:evaluation}

We conduct an evaluation to answer three research questions (RQs).
This section describes the evaluation metrics, datasets, and candidate models included in the evalution.

\noindent\textbf{\RQOneAlt}

\noindent\textbf{\RQTwo}

\noindent\textbf{\RQThree}

\subsection{Evaluation Datasets}
\label{sec:datasets}
\begin{table}[]
\caption{Details of the evaluation datasets.
Concept drift is abbreviated as CDs.
}
    \centering
    \scalebox{1}{
    \begin{tabular}{p{0.18\columnwidth}p{0.1\columnwidth}p{0.08\columnwidth}p{0.44\columnwidth}}
    \toprule
        Dataset (Source)               & Shape & \# CDs & Description\\\midrule
        \textbf{RQ1 \& RQ2}\\ \midrule
         ROUTER (proprietary) & 46,694 x 32  & 2 & internal dataset predicting failures\\
         Sine~\cite{gama2004}
                              & 16,000 x 2  & 2 & synthetic dataset\\
         SEA~\cite{dataset_sea}
                              & 16,000 x 2  & 2 & synthetic dataset\\
         ELEC~\cite{gama2004}
                              & 45,312 x 9  & 22 & real dataset predicting electricity price\\
         NOAA~\cite{6235959}
                              & 18,159 x 8  & 3 & real dataset predicting rain\\
         GCT~\cite{wilkes2020google}
                              & 263,509 x 19  & 28 & real dataset predicting job failure\\
         \midrule
         \textbf{RQ3} \\
         \midrule
         FM~\cite{xiao2017/online}
                              & 70,000 x 784  & 1 & real dataset predicting clothing type\\
         LMR~\cite{maas-EtAl:2011:ACL-HLT2011}
                              & 50,000 x 250  & 1 & real dataset predicting sentiment\\
                  \bottomrule
    \end{tabular}
    }
    \label{tab:datasets}
\end{table}
We select eight datasets for our evaluation (Table~\ref{tab:datasets}).
Seven of which are publicly available collected from previous works on concept drift detection~\cite{gama2004, Lyu2023OnTM} which consisted of three synthetic datasets (Sine, SEA, FM) and four real-world datasets (ELEC, NOAA, LMR, GCT). Also, we include one internal, real-world run-time dataset collected from the deployment of congestion prediction at a telecommunication company, i.e., ROUTER.
In our use case, the datasets can be divided into two groups: with ground-truth (GT) and without GT concept drift. The following datasets contain known concept drifts: Sine, SEA, FM, LMR, and ROUTER. 
Of these datasets, Sine, SEA, and ROUTER 100\% do not have data drift. Details of the dataset are as follows:

\pa{ROUTER}: An industrial and proprietary network traffic dataset containing router data from hundreds of sites over one month. This dataset includes tens of features with two features representing the delay and stability of the packet traffic having the most impact on the final class. The labels signal whether a network congestion occurs in the next hour. This dataset contains two concept drifts, where the delay and stability signal the concept drift changes due to equipment upgrades. This is an extremely imbalanced dataset where even after a 10:1 down-sampling, the positive class still consists of fewer than 4.1\% of the data.

\pa{Sine}~\cite{gama2004}: A balanced, noise-free synthetic dataset with randomly generated features. 
For the original concept, the label is positive if the point lies below $y = sin(x)$, otherwise, it is negative. 
Two concept drifts are introduced in this dataset. The first is at 3,000 where the concept is the reverse of the original concept.
The second is at 10,000 where the label function is changed to $y = 0.5 + 0.3sin(3\pi x)$.

\pa{SEA}~\cite{dataset_sea}: The setup is similar to~\textbf{Sine}. For the original concept, the class is positive if the point has $att1 + att2 \geq 8$. Otherwise, the class is negative. The first concept drift changes the positive class to points with features $att1 + att2 \geq 9$ and the second concept drift changes it to points where $att1 + att2 \geq 7$.


\pa{Electricity} (ELEC)~\cite{gama2004}: This dataset includes data points representing 30-minute demand data from various locations, covering two months between 1996 and 1998. The label indicates the price change relative to the moving average of the past 24 hours. Data points representing price increases comprised 42.5\% of the dataset.


\pa{NOAA}~\cite{6235959}: A subset of the original NOAA dataset containing 50 years of weather data from Offutt Air Force Base in Bellevue, Nebraska.
The label represents whether rain was observed on each day, where rainy days comprised 31.4\% of the data.

\pa{Google cluster trace} (GCT)~\cite{Lyu2023OnTM}: This dataset contains the cleaned and under-sampled version of the original production cluster trace data released by Google in 2011~\cite{wilkes2020google}. The features represent the first five minutes of monitored job data, while the labels indicate whether the job ultimately failed. This dataset is extremely imbalanced. The non-failure class consisted of less than 3.2\% of the dataset even after a 3:1 down-sampling.

\pa{Fashion-MNIST} (FM)~\cite{xiao2017/online}: A dataset contains images showing individual articles of clothing at low resolution (28 by 28 pixels). We reduce the number of classes to two. If the class is 0, 2, 3, 4, 6 it is a Top and the label is 0. Otherwise, it is 1. After reduction, this is a balanced dataset.
We create a synthetic concept drift at 4,000 as follows.
From index 4,000, if the point's original classes are 4 or 6 (now both are 0), it is switched to 1.
In this case, it roughly translated to 366 items with their labels changed.
The concept drift detection is performed on the 10,000 images in the testing dataset.

\pa{Large Movie Review} (LMR)~\cite{maas-EtAl:2011:ACL-HLT2011}: A balanced movie reviews dataset from the Internet Movie Database. We introduce concept drift at index 2,000 where the classes are reversed.

\subsection{Subject ML Models Trained Using the Datasets}
\label{section:online_models}
In RQ1 and RQ2, the online models are random forest classifiers.
For RQ3, two models are trained: a three-layer NN for the FM dataset and a five-layer NN for the LMR dataset. 

We train the online models in RQ1 and RQ2 using the random forest classifiers implementation in scikit-learn~\cite{scikit-learn}.
The training data consists of the first 500/1,000 data points of the corresponding datasets including labels.
The remainder of the datasets are converted into a time series without labels for testing. Hence, all experiments are conducted online where concept drift detection is active for the entire time series.


In RQ3, we evaluate the generality of \tool, i.e., to what extent \toolS can detect concept drifts in different ML models as a model-agnostic approach. 
We trained two NNs implemented by Keras Sequential models~\cite{chollet2015keras, tensorflow2015-whitepaper} using the code and data from TensorFlow~\cite{tf_image_classification, tf_text_classification}. 
The first NN contains three layers for classifications of clothing images in the \textbf{Fashion-MNIST} dataset. The first layer flattens the input. Then, the second layer (128 nodes) learns the clothing features. Lastly, the third layer (10 nodes) outputs the classification results. 
This neural network is trained on the first 60,000 data points of this FM dataset. 
The second NN consists of five layers for sentimental classification of the textual \textbf{LMR} dataset. The first layer takes the integer-encoded text and looks up an embedding vector for each word. The third layer converts the sequence to a fixed length.
The last layer is densely connected with a single output node.
The second and fourth rounds are drop-out layers to reduce overfitting.
This neural network is trained on the first 25,000 data points of this LMR dataset.

For all RQs, we train the inspector models using the random forest algorithm~\cite{scikit-learn}. The inspector models are first trained offline using $w$ latest data points with labels in the training dataset. Then it is retrained online as described.

\subsection{Evaluation Metrics}
We use four metrics to evaluate the performance of each concept-drift detection technique. For each CDD, we report each metric as the average of five rounds using random seed. The details of the metrics are as follows.

\indent\textbf{Model accuracy (MAcc.) (\%)} is the accuracy of the classification task using the online ML model with a CDD. This is computed as the percentage of correct predictions among the total data points. 

\indent\textbf{Precision (\%)} is the percentage of true positives of all the detected concept drifts by a CDD. \textbf{Recall (\%)} is the percentage of true positives detected by a CDD among all the GT concept drifts. 
\begin{itemize}
    \item True positive: The number of concept drifts that are \textit{correctly} detected by one CDD. 
    \item False positive: The number of concept drifts detected by one CDD that are incorrect and do not match with the GT concept drifts.
    \item False negative: The number of GT concept drifts that a CDD fails to detect. 
\end{itemize}

\indent\textbf{Labels used (Lbl) (\%)} is the percentage of the data points that a CDD requests manual labels for.

\pa{Calculating the precision and recall when GT concept drift is unavailable}.
In this case, we use the result of PHT, the supervised CDD on the predictions of the original online model (no retrain) as the GT. While the concept drifts detected this way are not perfect, previous results show that they have high precision. Thus, it can reliably indicate where a subset of concept drifts are for comparison.

\section{Evaluation}
\label{sec:eval_results}
This section provides the details for each RQ, including the motivation, method, and results.
\subsection*{\RQOneAlt}
\pa{Motivation.}
Motivated by our industry needs, we want to examine the performance of the SOTA concept drift detection techniques on both open-source and proprietary datasets. 
We want to learn what their limitations are, what we can learn from them, and whether we can use the insights to devise better ideas for concept drift detection.
First, we want to learn how they perform on core functionality such as concept drift detection and the resulting accuracy.
Then we want to learn how they perform on practical concerns of the CDDs such as the false positive rate and the amount of labels needed.

\pa{Method.}
We answer this RQ by evaluating a supervised CDD (i.e., PHT~\cite{pagehinkley1954}) and a semi-supervised CDD (i.e., ECHO~\cite{Haque2016ECHO}) on the first six datasets in Table~\ref{tab:datasets}.
ECHO is a well-known SOTA semi-supervised CDD that improves upon its predecessor~\cite{Haque2016SAND}.
We run its official Java implementation~\cite{echoimplement} using their default configuration, only changing the warm-up window size (to~\texttt{-S 1000}). The default parameters are ideally their best configuration, and this change enables ECHO to cover at least 70\% of the data points of each dataset.
This choice comes after experimenting with various configurations, which end up with ECHO crashing when less than 30\% of datasets are executed, trivializing the comparison effort. We provide the true labels for all of ECHO's label requests.
We confirmed that the results are at least on par with that observed in ECHO's papers~\cite{Haque2016ECHO, Haque2016SAND} or works using them as baselines~\cite{Pinage2019ADD, Sriwatanasakdi2017ConceptDD} on equivalent datasets.
PHT is a supervised CDD commonly used in previous work~\cite{Gama2014ASO, Sethi2017OnTR, Cerqueira2021STUDDAS}. PHT is a variant of the Cumulative Sum~\cite{pagehinkley1954}, another well-known CDD used as the basis for concept drift detection in ECHO.
For our experiment, we use the PHT implementation in scikit-multiflow~\cite{skmultiflow}.
As we use PHT as a supervised CDD, we allow its access to 100\% of the GT labels in the test datasets.

These methods are evaluated on the datasets described in Section~\ref{sec:datasets}. As described in Section~\ref{section:online_models}, to evaluate PHT, we train a random forest classifier on the first 1,000 data points of the dataset with true labels to create the base model. The rest of the dataset is the test dataset. Once concept drift is detected, the base model is retrained using the most recent 1,000 data points with true labels.
\begin{table*}[]
    \centering
    \caption{The performance of three CDDs: PHT (supervised), ECHO (semi-supervised), and \toolS (semi-supervised).}
    \scalebox{1}{
    \begin{tabular}{c|r|rrrr|rrrr?rrrr}
    \toprule
        & MAcc. & \multicolumn{4}{c}{\textbf{PHT (supervised)}} & \multicolumn{4}{c}{\textbf{ECHO (semi-supervised)}} & \multicolumn{4}{c}{\textbf{\toolS (semi-supervised)}} \\
         &  w/o CDD& MAcc. & Prec. & recall & Lbl. 
          & MAcc.  & Prec. & recall & Lbl. 
          & MAcc.& Prec. & recall & Lbl.  \\\midrule
       ROUTER & 96.56&97.88 &  100.0&50.0 &100.0 
       & 94.60 & 14.3 & 100.0 &48.1
       & 96.76  & 71.4&100.0 & 0.6\\\midrule
       SEA  &  95.91& 95.93  &100.0 & 100.0 & 100.0
       & 86.44  & 50.0 & 50.0 & 100.0
       & 95.55  & 61.5 & 80.0 & 0.6\\
       Sine & 73.69& 73.63  & 100.0 & 50.0 & 100.0
       &62.25 & 100.0 & 50.0 & 0.03
       & 75.97 & 48.1&80.0 &0.6\\\midrule
       ELEC & 66.06   & 74.19  & 58.3     &63.6  & 100.0
       &71.50 & 10.9 & 81.8 & 48.73
       &77.30&30.4&72.7&0.6\\
       NOAA & 76.89 & 77.37  &66.6     &66.6  & 100.0
       & 77.70  & 6.5 & 100.0 & 100.0
       &77.51 & 15.0 & 80.0 & 0.6\\
       GCT & 96.91& 97.37 & 69.2 & 53.8 & 100.0 & 96.74& 1.3 & 75.0& 91.7&97.33 & 17.6 & 35.7 &0.8\\
       \bottomrule
    \end{tabular}
    }
    \label{tab:rq1_results}
\end{table*}

\pa{Results.} Table~\ref{tab:rq1_results} shows the results of applying PHT and ECHO for concept drift detection across six datasets.
We also show the model accuracy of the original online model on the datasets for comparison. It is expected that with the help of a CDD, the model accuracy will increase. Also, it is expected that the supervised CDD PHT outperforms the semi-supervised CDD ECHO.

Our evaluation results show that PHT and ECHO can improve the overall model accuracy, i.e., PHT contributes to a higher model accuracy for five of the six datasets and ECHO leads to a higher or comparable model accuracy for two of the six datasets. 
The limited improvements of PHT and ECHO are expected for two reasons. First, the datasets contain a relatively small number of concept drifts that span a small percentage of data points. Second, these datasets contain class imbalance, hence, wrong classifications for minority classes are not as impactful. 
In such cases, adapting to the new concepts could cause slight decreases in accuracy.
In practice, online models that process streaming data accumulated over weeks or months would greatly benefit from concept drift detection.


In terms of concept drift detection, our evaluation results confirm that the supervised CDD PHT achieves 100\% precision in the three datasets with GT concept drifts, i.e., ROUTER, SEA, and Sine. PHT fails to detect some concept drifts in ROUTER and Sine, i.e., the recall is 50\%. A lower recall is expected considering the conservative design of PHT, which favors precision over recall. 
As a semi-supervised CDD, ECHO's precision is lower and more unstable (ranging from 1.3\% to 100\%) on the six datasets compared to PHT. 
The recall of ECHO is comparable to that of PHT, ranging from 50\% to 100\%. 

Another important metric of CDDs is the amount of GT labels required, i.e., expert labeling needs to be performed manually. A highly practical CDD is expected to have a low percentage of manual labels required. 
As PHT is supervised, it requires 100\% of the data to be manually labeled. 
The semi-supervised CDD ECHO requires significantly fewer labels.
However, our evaluation results find that the percentage of labels required by ECHO is unstable with a large range across different datasets, as low as 0.03\% for Sine, yet an excessive amount of labeling is required for the other datasets (48.1\% to 100\%).




\noindent\textbf{Discussions on using PHT and ECHO in industry settings.} 
Our evaluation of PHT and ECHO exhibits significant inconsistency and raises concerns from industry partners as we cannot foresee whether the target dataset would require a lot of true labels.
In our observation, the number of labels required for SOTA methods varies a lot between datasets which allows for our consistency to stand out for planning the labeling efforts.
Also, as industrial partners already have pre-trained models to solve their problems, switching to another architecture with unknown results is not desirable.
While PHT can work with existing models, it requires 100\% data points to be labeled which is unrealistic. On the other hand, ECHO utilizes an ensemble of centroid-based k-NN models for classification and concept drift detection~\cite{Haque2016ECHO}. This type of model is not necessarily adaptable to all use cases.

\rqboxc{Current SOTA either demands impractically large amounts of labeled data or is incompatible with pre-trained industry models.}

\subsection*{\RQTwo}
\pa{Motivation.}
Since the SOTA CDDs do not satisfy industry needs, we propose \toolS including two key innovations, i.e., clustering data points and spreading true labels to reduce the excessive labeling required while improving model accuracy.
In this RQ, we want to examine the performance of \toolS in two steps.
First, we evaluate the performance of \toolS in concept drift detection using the default configuration. 
Second, we want to examine the performance of \toolS under different window and memory sizes. These two parameters are configurable and significantly influence \tool's effectiveness in concept drift detection. The window determines which data points are labeled and stored in the memory, while the memory determines the training data labels for the inspector model.

\pa{Method.}

\pa{\underline{- Using \tool's default configuration.}} We run \toolS on the first six datasets in Table~\ref{tab:datasets}.
For each dataset, we use the first 1,000 data points to train a random forest classifier to be the online model.
In this part, we set the windows and memory sizes to our default configuration. The default window size is 1,000 and the default memory size is 15.

\pa{\underline{- Using different configurations of \tool.}} 
We run \toolS on the three datasets with known concept drifts, i.e., ROUTER, SEA, and Sine. 
We vary the window and the memory size used to train the inspector model and observe the resulting impact on performance.
We experiment with memory with queue sizes 10 and 15.
For the window size, we try two values of 500 and 1,000, where the latter is the same size as the training data used to train the online models. 
Each pair of memory and window size will affect the retraining process of the inspector model, changing its prediction and the resulting concept drift detected.

For both steps, we use PHT as the integrated supervised CDD to detect changes in the prediction error distribution between the original and the inspector models (Algorithm~\ref{alg:framework_algo}).
Similar to RQ1, we provide the CDD with all labels requested and evaluate model accuracy, precision, recall, and percentage of labels required.

\pa{Results.}

\underline{\pa{- The performance of \tool.}}
Table~\ref{tab:rq1_results} shows the results of \toolS side-by-side to PHT and ECHO. Our evaluation shows that \toolS improves the accuracy of the online models for all five datasets except for one, i.e., SEA. \toolS has comparable performance with the supervised PHT, with slightly higher model accuracy for three of the six datasets. Compared to the semi-supervised ECHO,~\toolS achieves higher model accuracy on all six datasets. \textit{In short, our \toolS leads to comparable performance improvement on model accuracy with the supervised PHT and outperforms the SOTA semi-supervised ECHO.}




As for concept drift detection, \toolS outperforms ECHO on both precision and recall and outperforms PHT in recall but has worse precision.
For the three datasets with GT labels, \toolS has acceptable precision (more than 58.9\%) and high recall (more than 90\%). 
For the three datasets with PHT-no-retraining GT labels, \toolS has much higher precision and lower recall than ECHO. This is due to ECHO generating significantly more (false) alarms.
\toolS has lower precision than PHT, which is expected as PHT has access to all labels.  


Concerning the manual labeling needed (`Lbl' in Table~\ref{tab:rq1_results}), \toolS requires between 0.6\% and 0.8\% of the data points, which corresponds to the number of data clusters as described in section~\ref{sec:method_overview}. 
This shows that \tool's label requirement is the lowest for all datasets except Sine. Our investigation reveals that ECHO's 0.03\% label requirement on the Sine dataset stems from a shortage of data points with low estimated confidence, leading to a very limited number of points that meet the selection criteria. However, Sine is an exception; for the remaining datasets, ECHO requires at least 48.1\% of the data for manual labeling, which is impractical.
Furthermore, \toolS requests a stable range of data for manual labeling, in contrast to the large range of data required by ECHO.
The count of the labeling requests (which may request for one or more labels each) shows an even larger gap.
For example, the ROUTER dataset represents one month's worth of data, where 800 data points are logged each hour. During this period, ECHO makes 22,459 requests for data labels, while~\toolS makes 80 requests. Considering the number of labels, the number of requests, and the overhead associated with processing each request, this makes a significant difference to the experts' labeling efforts.
Overall, \toolS shows detection performance competitive to the state-of-the-art while utilizing fewer labeled data points. Depending on the dataset, our method significantly reduces the number of labels used while maintaining comparable detection results and accuracy.

\begin{table}[!htbp]
\caption{The performance of \toolS under different configurations of window size and memory size.}
\centering
\scalebox{1}{
\begin{tabular}{lllllll}
\toprule
Dataset                
    & Window & Memory & Precision        & Recall        & Lbl.     \\
    & Size & Size & & &\\
\toprule
\multirow{4}{*}{ROUTER} 
    & 500   & 10    & 71.4           & 100.0     & 0.9 \\
    & 500   & 15    & 100.0          & 60.0     & 0.7 \\
    & 1000   &10    & 57.1           & 80.0   & 0.8 \\
& \textbf{1000}  & \textbf{15}    & \textbf{71.4}  & \textbf{100.0}  & \textbf{0.6} \\ 

\midrule
\multirow{4}{*}{Sine}  
    & \textbf{500}   & \textbf{10}    & \textbf{58.8}   & \textbf{100.0}     & \textbf{0.6}     \\
    & 500   & 15    & 50.0            & 70.0       & 0.6 \\
    & 1000  & 10    & 52.6            & 100.0      & 0.7 \\
    & 1000  & 15    & 48.1            & 80.0       & 0.6 \\
    \midrule
\multirow{4}{*}{SEA}    
    & \textbf{500}  & \textbf{10}  & \textbf{75.0}  & \textbf{90.0}     & \textbf{0.7} \\
    & 500  & 15  & 80.0           & 80.0              & 0.7 \\
    & 1000 & 10  & 41.7           & 50.0       & 0.6 \\
   & 1000 & 15   & 61.5           & 80.0     & 0.6 \\

    \bottomrule
\end{tabular}
}
\label{tab:rq1_window_size}
\end{table}

\underline{\pa{- Sensitivity of \tool's performance.}} Table~\ref{tab:rq1_window_size} shows the performance of \toolS with different configurations. First, we look at how window size impacts our detection results.
Our results show that the best window size varies for different datasets.
The window size of 1,000 works best for the ROUTER dataset while the window size of 500 is best for Sine and SEA. 
Similar to the window size, the best-performance memory size varies for each dataset.
We observe that a larger memory size may sacrifice recall for precision. Hence, memory size should be adjusted according to the size of the concept drift we want \toolS to capture. For concept drifts that occur only briefly, memory size should be smaller and vice versa.
However, this might lead to decreased precision. 
Lastly, our evaluation shows that the percentage of manual labeling required by \toolS is stable for different combinations of window and memory sizes. 
Labeling efforts can be further reduced at the cost of lower precision, presenting opportunities to label fewer points than requested if desired.
\textit{Note that with the default configuration, \toolS still outperforms the SOTA. Our experiments confirm that \toolS can be tuned to have an even better performance for each dataset.}

\noindent\textbf{Discussions on adopting \toolS in industry settings.} We discuss the adoption of \toolS with our industry partners and conclude that the precision and recall of \toolS are sufficient for deployment in the industry. 
In practice, higher recall is preferred to be more alert to potential concept drifts, as they can be quickly verified to rule out false alarms.
In addition,~\tool's extremely low manual labeling effort, especially in terms of the number of requests, is desirable for planning and reducing labeling efforts. From a practical standpoint, the saved effort can be redirected to examine reported concept drift.
Couple that with addressing experts' preference of labeling multiple data points at their convenience and the flexibility in the points to be labeled makes~\toolS suitable for adoption.


\rqboxc{\toolS is on par with supervised CDD in terms of the accuracy of online models and outperforms the SOTA semi-supervised ECHO. \toolS outperforms ECHO for detecting concept drifts with higher precision and recall, yet requiring significantly less manual effort for labeling. 
\toolS addresses multiple industry concerns about the practicality and performance of concept drift detection, allowing deployment in industrial settings.
}

\subsection*{\RQThree}

\pa{Motivation.}
We followed the convention of datasets and model training in previous RQs. Yet, it remains unclear whether \toolS can generalize to different ML models. 
Also, as industrial partners possess pre-trained models for their use cases, switching to other untested architecture is undesirable.
While PHT can work with existing models, it requires 100\% data points to be labeled which is unrealistic. On the other hand, ECHO utilizes an ensemble of centroid-based k-NN models for classification and concept drift detection~\cite{Haque2016ECHO} making it not necessarily adaptable to other use cases.
Hence in this RQ, we examine \tool's performance on pre-trained classifiers.

\pa{Method.}
We include two different Keras sequential neural networks for image classifications. The first model consists of three layers and the second model consists of five layers as described in Section~\ref{section:online_models}. We trained the two models using the released data and source code from Tensorflow document~\cite{tf_image_classification, tf_text_classification}. Model 1 is trained using the complete 60,000 data points from the FM dataset and Model 2 is trained using the 25,000 points from the LMR dataset.

To simulate the production application, we train a random forest classifier using the last 1,000 points of the training data as the inspector model. 
Benefiting from more than one order of magnitude smaller training set, the inspector model is trained using significantly less time than the original model.
In this RQ, we configure \toolS to use the window size of 1,000 and the memory size of 20. Similar to prior RQs, considering the randomness of \tool, we run the experiments for five runs and calculate an average precision and recall.

\pa{Results.}
Our evaluation shows that \toolS achieves a precision of 62.5\% and a recall of 100\% for Model 1 (an image classification neural network), and a precision of 80\% and a recall of 80\% for Model 2 (a text classification neural network). The performance of \toolS on alternative ML models is similar to that in RQ2. As for the percentage of labels required for manual labeling, consistent with RQ2, we find \toolS requires 0.99\% and 0.98\% for the two models respectively, which are around 300 to 600 data points.



\rqboxc{\toolS successfully detects concept drift for pre-trained models and is not limited by a model architecture (unlike ECHO) while requiring little manual labeling.}

\section{Threats to Validity}
\pa{Threats to internal validity.} \toolS has parameters that may impact its performance. Our sensitivity experiment to assess the impact of different parameters on \tool's performance shows that \tool's memory size should be proportional to the size of the concept drift for best recall. However, the default values are sufficient for most cases.



\pa{Threats to construct validity.} We describe the threats below regarding propagation and the percentage of labels required.
First, the learning method to propagate the labeled data points to unlabeled data points can change \tool's performance. While we observed similar results from $Label Spreading$ and $Label Propagation$ models from sklearn~\cite{scikit-learn}, future works can investigate more propagation methods.
Second, our method has inherent randomness from the sampling process. We expect a fluctuating number of labels required for manual labeling between runs.
To address this issue, we report an average number from five runs in this paper. 

\pa{Threats to external validity.} Our evaluation is limited to the datasets included in this paper, and may not work for every type of concept drift. We include the commonly used and popular datasets from prior works on concept drifts. In addition, we include one proprietary dataset from industry partners and one from Google. 
\toolS is model-agnostic by design. 
We show that \toolS works on three models with different architectures.
However, the performance of \toolS on other models remains unknown.

\section{Conclusion}
We propose \toolS, a concept drift detection framework that outperforms the state-of-the-art in precision and recall while requiring fewer manual labels and offering more flexibility in data distribution and model architecture. \toolS addresses industry challenges and integrates easily into existing maintenance frameworks. The implementation of CDSeer and the ROUTER dataset are close-sourced. However, we provide sufficient description to re-implement the proposed method. As for evaluation, seven of the described datasets are publicly available for reproduction.

\section*{Acknowledgment}
This work is supported by Mitacs and Ericsson (GAIA
Montreal, Canada).

\bibliographystyle{IEEEtran}
\bibliography{paper}
\end{document}